\documentclass[11pt]{article}
\usepackage[preprint]{acl}
\usepackage{times,latexsym}
\usepackage[T1]{fontenc}
\usepackage[utf8]{inputenc}
\usepackage{microtype,inconsolata,needspace}
\usepackage{amsmath,amssymb,booktabs,tabularx,multirow}
\usepackage{graphicx,tikz}
\usepackage{xcolor}
\usepackage{gradient-text}
\usepackage{xspace}
\DeclareRobustCommand{\model}{%
  \mbox{\gradientRGB{EviGraph}{92,122,255}{0,168,150}}\xspace}
\usetikzlibrary{arrows.meta,positioning,fit}
\newcommand{\recommend}{\textsc{Recommend}}
\newcommand{\verify}{\textsc{Verify}}
\newcommand{\exclude}{\textsc{Exclude}}
\newcommand{\BaseCount}{4,500}
\newcommand{\CompositionalCount}{300}
\newcommand{\InheritedChallengeCount}{150}
\newcommand{\StressCount}{1,500}
\newcommand{\AddedChallengeCount}{150}
\newcommand{\ScenarioCount}{6,300}
\newcommand{\ScoredCount}{25,500}
\newcommand{\InheritedPositiveCount}{10}

\newcommand{\DevRuleMeanFone}{0.865}
\newcommand{\DevUnverifiedMeanFone}{0.831}
\newcommand{\DevVerifiedMeanFone}{0.810}

\newcommand{\DevRuleNoTimeCompRisk}{0.571}
\newcommand{\DevRuleNoTimeStressRisk}{0.143}
\newcommand{\DevRuleNoScopeBaseFone}{1.000}

\newcommand{\BlindRuleBaseFone}{0.915}

\newcommand{\BlindRuleCompFone}{0.816}

\newcommand{\BlindRuleStressFone}{0.927}

\newcommand{\BlindVerifiedBaseFone}{0.900}

\newcommand{\BlindVerifiedBaseRecRecall}{0.869}

\newcommand{\BlindVerifiedCompFone}{0.864}

\newcommand{\BlindVerifiedStressFone}{0.900}

\newcommand{\BlindVerifiedStressRecRecall}{0.880}

\newcommand{\BlindInitialRecommendations}{238}
\newcommand{\BlindFinalRecommendations}{217}
\newcommand{\BlindSupportedWithdrawals}{21}
\newcommand{\BlindUnsupportedWithdrawals}{0}
\newcommand{\BlindAdjudicatorCalls}{31}
\newcommand{\BlindAdjudicationRestorations}{10}
\newcommand{\StabilityWorstBaseRecall}{0.688}
\newcommand{\StabilityMeanBaseRecall}{0.833}
\newcommand{\StabilitySDBaseRecall}{0.115}

\newcommand{\ExtOpusVerifiedBaseFone}{0.944}

\newcommand{\ExtSonnetVerifiedBaseFone}{0.902}

\title{\model{}: Proof-Carrying Selective Recommendation over\\
Temporal Public-Service Knowledge Graphs}

\author{
\textbf{Yixi Zhou}\textsuperscript{1}\thanks{Equal contribution} \
\textbf{Sikun Wang}\textsuperscript{2}\footnotemark[1]\
\textbf{Lei Fan}\textsuperscript{3}\
\textbf{Fan Zhang}\textsuperscript{4}\thanks{Corresponding author} \\
\textsuperscript{1}Hong Kong Baptist University \
\textsuperscript{2}Tokyo University of Science \\
\textsuperscript{3}University of Illinois Urbana-Champaign  \
\textsuperscript{4}The University of Tokyo \\
yxzhou@comp.hkbu.edu.hk, jc19546883@gmail.com  \\
leifan2@illnois.edu, zhang-fan@g.ecc.u-tokyo.ac.jp
}

\begin{document}
\maketitle
\begin{abstract}
Public-service recommendations require evidence that matches the requested
service, scope, and date. Yet treating every missing detail as decisive can
withhold useful recommendations. We introduce \model{}, which distinguishes
critical decision requirements from information that can remain unresolved.
A language agent links these requirements to evidence in a temporal knowledge
graph, while a deterministic checker establishes whether a recommendation is
supported. Evaluation on a bilingual Hong Kong public-service benchmark with
executable policy references shows that this distinction reduces unnecessary
abstention. Additional verification, however, can withdraw supported
recommendations without improving decision quality. These findings suggest
that reliable evidence-based navigation depends on specifying what must be
established for a decision, rather than simply adding more verification.
\end{abstract}
\section{Introduction}
A family seeking occasional child care next month may find current kindergarten
vacancies at the same institution. This relevant evidence establishes neither
occasional-care availability nor a future booking. Conversely, treating every
missing administrative detail as decisive can withhold a supported
recommendation. Useful navigation requires distinguishing decision-critical
evidence from information that belongs in a follow-up step.

Graph-guided retrieval connects evidence across entities and documents
\citep{sun2024tog,ma2025tog2,zhu-etal-2025-knowledge}.
Knowledge graphs (KGs) retain a claim's service, scope, time, and provenance,
but do not themselves determine which paths justify a recommendation.
We study the evidence obligations that govern selective public-service
decisions and evaluate the contribution of additional verification to decision
quality when the supporting proof has already been checked.

\model{} makes these obligations explicit. Hard eligibility and essential
recommendation evidence are critical; procedures and preferences remain visible
without independently blocking a recommendation. A large language model (LLM)
interprets the request and proposes candidate-local evidence links. A
deterministic compiler checks their applicability and retains a proof.
A bounded verifier can challenge existing critical paths, followed by at most
one adjudication. This interface separates verification from proof construction.


Our contributions are:
\begin{itemize}
\item \textbf{A typed decision interface.} We distinguish unresolved evidence
from demonstrated ineligibility and make the conditions of a recommendation
inspectable through obligation witnesses.
\item \textbf{A controlled public-service benchmark.} Bilingual Hong Kong care
requests combine service, time, and provenance constraints. Disjoint entities
and structural templates support held-out comparisons, with executable policy
references that expose their assumptions.
\item \textbf{Evidence about verification trade-offs.} \model{} recovers recall
relative to an all-field compiler, but its verifier removes supported
recommendations without removing unsupported ones. A strong rule-only baseline
also limits claims of an intrinsic advantage from using a language agent.
These findings concern agreement with executable policy references; positive
recommendation evidence is concentrated in kindergarten navigation, without
establishing current service availability.
\end{itemize}

\clearpage
\section{Related Work}
\paragraph{Retrieval and knowledge graphs.}
Retrieval-augmented generation grounds language models in external memory
\citep{NEURIPS2020_6b493230}. FLARE retrieves during generation when anticipated
content is uncertain \citep{jiang-etal-2023-active}; Self-RAG learns when to
retrieve and how to critique retrieved evidence and generations
\citep{asai2024selfrag}. Irrelevant context can nevertheless reduce accuracy,
motivating entailment filtering and robustness training \citep{yoran2024robust}.
Think-on-Graph explores graph paths \citep{sun2024tog}, while ToG-2 couples
graph and document retrieval \citep{ma2025tog2}. KG$^2$RAG organizes evidence
through fact-level relations \citep{zhu-etal-2025-knowledge}, and K-RagRec applies
graph retrieval to recommendation \citep{wang-etal-2025-knowledge-graph}.
\model{} instead studies which evidence authorizes a decision over fixed
candidates, including service, scope, and time constraints.

\paragraph{Attribution and verification.}
ALCE evaluates answer quality and citation support
\citep{gao-etal-2023-enabling}; FActScore assesses support for individual atomic
facts \citep{min-etal-2023-factscore}. RARR retrieves evidence and revises
unsupported statements \citep{gao-etal-2023-rarr}, while CRITIC uses external
tool feedback to improve outputs \citep{gou2024critic}. Intrinsic
self-correction without external feedback can fail or degrade reasoning
\citep{huang2024selfcorrect}. Our verifier can inspect existing evidence, but
its criticisms must identify a decision-critical defect. We therefore measure
whether verification improves decisions, beyond whether citations are valid.

\paragraph{Selective prediction and uncertainty.}
SelectiveNet learns prediction and rejection \citep{pmlr-v97-geifman19a}, and
selective question answering studies coverage under domain shift
\citep{kamath-etal-2020-selective}. Semantic entropy groups equivalent answers
to estimate uncertainty \citep{kuhn2023semantic}. Conformal factuality reduces
output specificity to obtain calibrated correctness guarantees
\citep{pmlr-v235-mohri24a}. \model{} distinguishes demonstrated ineligibility
from insufficient evidence: the former permits exclusion, whereas the latter
calls for verification. Our confidence curves are descriptive and do not
provide conformal guarantees.

\paragraph{Structured reasoning and proof interfaces.}
FAME constructs entailment trees through planning
\citep{hong-etal-2023-faithful}; LINC translates natural-language premises into
first-order logic for an external prover \citep{olausson-etal-2023-linc}.
Proof-carrying code separates an untrusted producer from a policy checker
\citep{Necula97popl}. \model{} adopts that separation for typed obligation
witnesses over extracted facts. Its checks establish applicability under the
declared policy, conditional on query interpretation and extraction; they do
not certify real-world admission or availability.

\section{Task and Benchmark}
\begin{table*}[t]\centering\small\setlength{\tabcolsep}{3pt}
\begin{tabular}{llrrrr}\toprule
View & Group & Queries & Candidates & R/V/E & S/C/T/M\\\midrule
\multirow{3}{*}{Partition} & Base & 1,228 & 6,140 & 84/5393/663 & 110/0/0/6030\\
 & Compositional & 92 & 460 & 31/391/38 & 40/0/0/420\\
 & Stress & 606 & 606 & 133/441/32 & 178/166/55/207\\
\midrule
\multirow{3}{*}{Domain} & Child/family & 1,261 & 4,441 & 248/4141/52 & 328/122/55/3936\\
 & Disability & 276 & 1,212 & 0/892/320 & 0/0/0/1212\\
 & Elderly & 389 & 1,553 & 0/1192/361 & 0/44/0/1509\\
\midrule
\multirow{3}{*}{Language} & Bilingual & 1,319 & 4,951 & 158/4305/488 & 217/120/34/4580\\
 & English & 316 & 1,136 & 53/935/148 & 66/26/12/1032\\
 & Trad. Chinese & 291 & 1,119 & 37/985/97 & 45/20/9/1045\\
\bottomrule\end{tabular}
\caption{Held-out benchmark support, independent of model success. R/V/E denotes automated-policy RECOMMEND/VERIFY/EXCLUDE; S/C/T/M denotes SUPPORTED/CONFLICTING/STALE/MISSING. The three views describe the same cases and must not be summed together. Compositional includes current/future contrasts. Full validation and construction counts are in the supplementary material.}\label{tab:data}\end{table*}

\subsection{Decision interface}
Let $q$ denote a request, $t$ its evidence cutoff, and $S_q$ its candidate
service set. Each candidate has public claims, published rules, and
source-linked background assertions. The output specifies eligibility,
evidence state, and a selective decision. \recommend{} supports the navigation
need without reserving a place; \exclude{} requires demonstrated hard
ineligibility; \verify{} retains unresolved essential conditions.

Evidence states aggregate in priority order: \textsc{Conflicting},
\textsc{Missing}, \textsc{Stale}, then \textsc{Supported}. A published zero
vacancy can be supported information without justifying a recommendation.

\subsection{Sources and reference construction}
The benchmark contains \ScenarioCount{} scenarios across three partitions:
\BaseCount{} Base requests, \CompositionalCount{} Compositional requests, and
\StressCount{} Stress scenarios, yielding \ScoredCount{} candidate decisions.
All draw on historical public-service evidence. Claims retain their sources,
relevant dates, and service scope; general provider descriptions cannot
establish a vacancy or a person's eligibility.

Compositional requests combine multiple requirements and include
\AddedChallengeCount{} current/future contrasts across three domains. These
contrasts test temporal applicability while holding source facts fixed.
Kindergarten vacancies supply positive recommendation references; other care
domains retain unresolved provider-specific needs. Base and Compositional
evaluate every candidate, whereas Stress evaluates the perturbed target.
Table~\ref{tab:data} reports held-out support; the appendix details construction
and the concentration of positive examples in child/family services.

Reference labels apply explicit research-policy rules to each request and its
evidence. They are automated judgments, not human assessments or observed
service outcomes. Information used to construct the expected decisions is
withheld from model inputs. Controlled stress cases remove facts, change dates,
reverse evidence order, or introduce contradictory claims. These interventions
test whether decisions respond to relevant evidence changes rather than to
incidental presentation.

\subsection{Entity and template separation}
Entity-relation clusters belong to fit, calibration, internal validation, or
blind test. Structural template relations group domain, service family, and
rendered requirement fields, ignoring values, language, identifiers, and
neutral framing. Stress variants inherit their parent relation. This holds
out requirement compositions, without claiming unseen service families.

Candidate sets are formed without outcome labels. Source characteristics
determine admissible stress interventions. We select the first assignment
meeting prespecified support requirements before observing predictions.
Entities, services, and structural templates do not overlap across splits;
semantic similarities may remain. The supplement details sampling choices
and their effects on the evaluation distribution.

\section{\model{}}
\subsection{Typed evidence obligations}
\begin{figure*}[t]
\centering
\includegraphics[width=\linewidth]{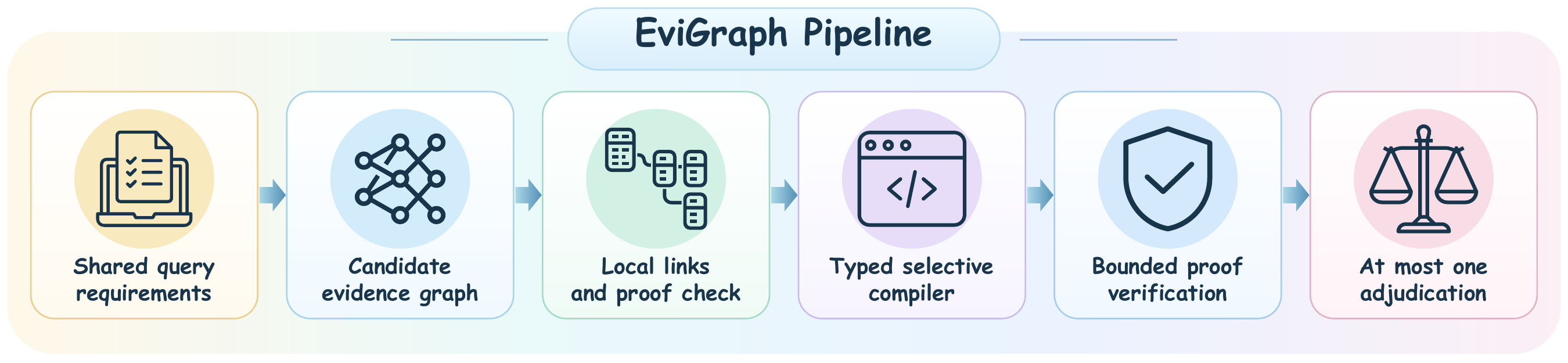}
\caption{Proposed \model{} pipeline. Query semantics are shared across candidates. Only an initial recommendation can enter verification or be restored by adjudication.}\label{fig:overview}
\end{figure*}

For request $q$ and candidate service $s$, let $G_{q,s}=(V,\mathcal{E})$
be the visible evidence graph, with node set $V$ and edge set $\mathcal{E}$.
Figure~\ref{fig:overview} separates language interpretation from deterministic
checks of proposed evidence paths.
Nodes represent query requirements, policy constraints, candidate claims,
services, scopes, time intervals, and sources. Typed edges retain applicability,
support, contradiction, scope, time, and provenance. Each evidence node points to an available source record.

Let $O(q,s)$ be the applicable obligations for candidate $s$. A typing function
$\tau$ assigns hard eligibility, essential recommendation, procedure, preference,
informational, or uncertain roles. Let $H$ and $E$ denote hard eligibility and essential recommendation roles,
respectively. The critical set is
\begin{equation}
C(q,s)=\{o\in O(q,s):\tau(o)\in\{H,E\}\},
\end{equation}
A rule's procedural axis takes precedence over administrative wording such as
\emph{hard}: required paperwork is not a proved person-side eligibility failure.

\subsection{Shared query extraction and local linking}
The agent extracts requirements once per scenario, shared across candidates.
Each stores a field, criticality, necessity, normalized value, scope, and exact
query quote. Unavailable quotes invalidate facts; conflicting values remain
unresolved. Requested grade and actual attendance grade are distinct.

The agent links the established obligations to visible rules or claims. It cannot change
query facts for a particular candidate. Deterministic checks reject evidence from other candidates,
nonexistent paths, incompatible service/grade/session scope, and invalid
time relationships. Published rule operators compare person facts; stated needs
cannot establish provider capabilities.

\subsection{Time and scope are proof conditions}
\begin{figure}[tbp]\centering
\includegraphics[width=\linewidth]{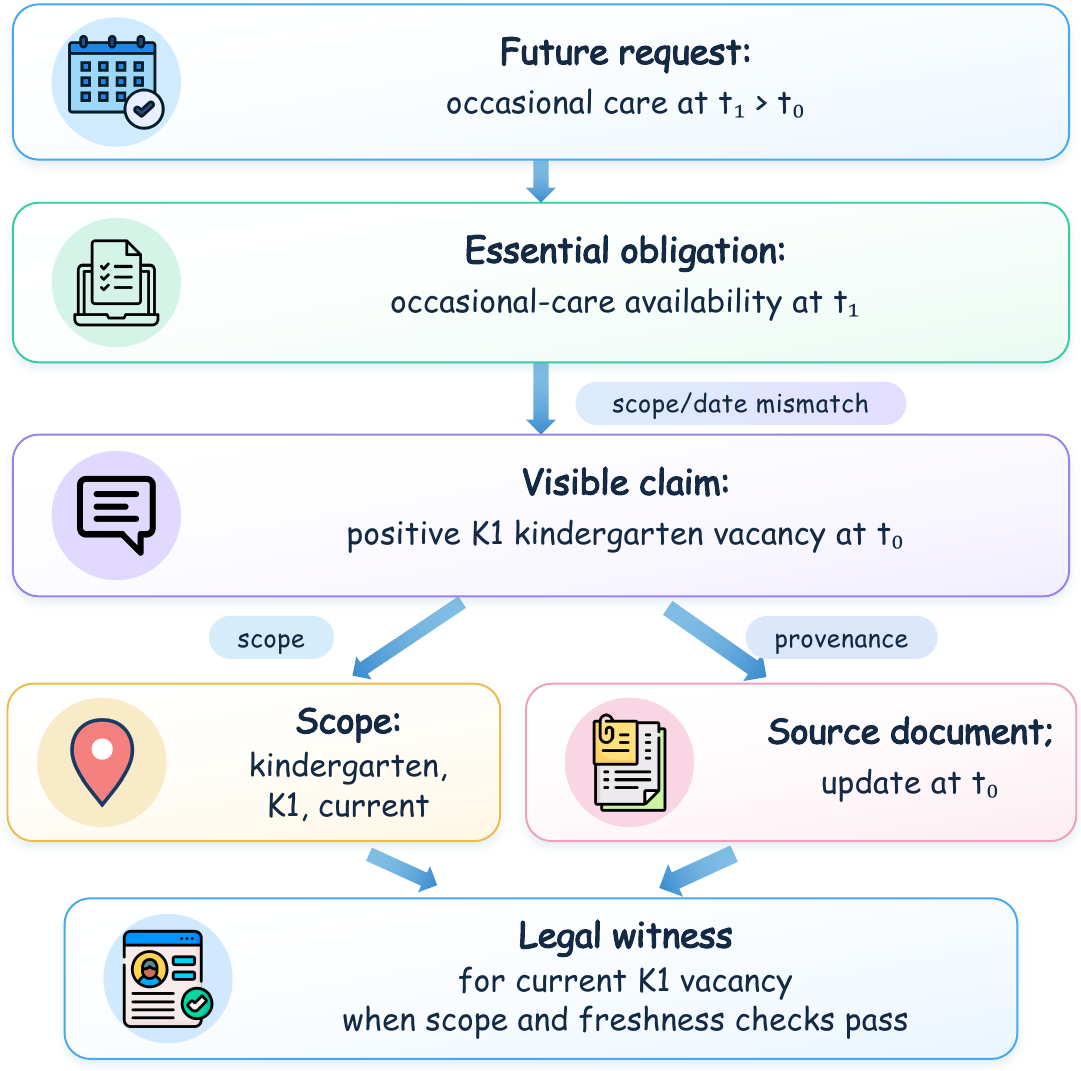}
\caption{Schematic proof conditions, not an observed prediction. Current kindergarten vacancy can provide a valid witness for that service and time; the scope and date mismatch prevents it from satisfying the future occasional-care obligation. A recommendation additionally requires every other critical obligation.}\label{fig:scope}\end{figure}

A kindergarten vacancy cannot establish occasional-care or future-date
availability. Observation and explicit validity edges are distinct; retrieval
time cannot substitute for an absent update timestamp. Dynamic claims use
prespecified freshness windows for each source. The checker compares
observations with compatible scope and effective dates before assessing freshness. Same-scope, same-time conflicts persist
regardless of evidence order. Figure~\ref{fig:scope} separates a supported
current fact from an unsupported recommendation path.

\subsection{Minimal sufficient proof and selective compilation}
For obligation $o$ and proof $\pi$, let $\operatorname{pass}(o,\pi)$ mean that $\pi$ supplies a valid witness
for obligation $o$. A sufficient positive proof satisfies
\begin{equation}
\forall o\in C(q,s),\quad \operatorname{pass}(o,\pi)=1.
\end{equation}
The certificate retains one representative of equivalent supporting claims
and each distinct required rule condition. It is inclusion-minimal within
this witness representation, not a globally minimum proof.

Let $O_H=\{o\in O(q,s):\tau(o)=H\}$ denote hard-eligibility
obligations and write $C=C(q,s)$. Let $\operatorname{fail}(o)$ denote
a demonstrated violation. Suppressing the proof argument, the compiler's
decision $D(q,s)$ is
\begin{equation}
D(q,s)=\begin{cases}
\exclude, & \exists o\in O_H:\operatorname{fail}(o),\\
\recommend, & \forall o\in C:\operatorname{pass}(o),\\
\verify, & \text{otherwise}.
\end{cases}
\end{equation}
The positive branch also requires supported essential evidence and a nonempty
essential obligation set. Procedural and preference gaps remain in the
explanation; they cannot become critical blockers merely because they are
unresolved.

\paragraph{Policy-relative guarantee.}
Given correct extraction, complete obligations, an accepted research policy,
and a correct checker, the positive branch guarantees a witness for every
critical obligation. This guarantees neither real-world availability nor
correct language interpretation.

\paragraph{Noncritical invariance.}
Holding $C(q,s)$ and its proofs fixed, changing an obligation outside $C(q,s)$
cannot change the compiler's decision, although explanations and confidence can change.

\subsection{Bounded verification and adjudication}
Only initial recommendations reach the verifier. Downgrades must cite an
existing critical obligation, retained proof edge, candidate evidence, and
scope or time defect; other concerns are rejected. One adjudication may
restore an original recommendation by answering every accepted blocker with
valid paths. Initial \verify{} and \exclude{} decisions cannot be upgraded.
All calls and accepted or rejected blockers remain logged.

\clearpage
\section{Experimental Setup}
\subsection{Models and comparisons}
We evaluate the effect of typed evidence obligations and verification using
GPT-5-mini as the primary language-model backbone. All methods operate on the
same candidate set and underlying evidence pool, with different retrieval,
representation, and processing designs. The appendix provides execution settings.

We compare Rule-only execution, Direct prediction from flat evidence,
Flat RAG using BM25 retrieval, Critic-and-guard, an All-field compiler baseline,
\model{} w/o verifier, and full \model{}.
Rule-only parses visible requests without privileged construction information.
Flat RAG selects up to twelve claims per candidate and retains all published rules.

The fixed internal-validation subset contains one hundred base, fifty
compositional and one hundred stress scenarios. Stratification by domain, language, and stress type ensures sufficient
positive examples for evaluation.
This support-enriched development instrument does not estimate natural request
prevalence. Methods, prompts, evidence roles, and decision thresholds are fixed
before test evaluation. The appendix details validation and retry rules.

\paragraph{Exploratory cross-backbone evaluation.}
We also compare Direct, Flat RAG, and \model{} with and without verification
using GPT-4o mini and Claude 4.5 on the same evaluation data. This analysis was
designed after accessing the primary GPT-5-mini test results and is not an
independent blind confirmation. Claude results cover all configurations;
GPT-4o mini contributes only complete method--partition units, detailed in
the appendix.

\subsection{Ablations and stability}
We remove criticality distinctions, scope checks, time checks, semantic
proof checks, or the verifier, and separately make every information gap
blocking. The verifier ablation is \model{} w/o verifier, not an independent
replicate. Compiler ablations share archived query extractions and link
proposals to isolate the changed check.

We assess \model{}'s stability over three independent development runs,
reporting the mean, population standard deviation, and worst observed result.
Repeated evaluation of the deterministic baseline checks reproducibility.
Baseline selection and experimental settings are fixed using development
data before evaluating the held-out test partitions.

\subsection{Evaluation measures and statistical tests}
We use fixed-label decision macro-F1 to give each of the three decisions equal
weight despite class imbalance. Present-reference
macro-F1 excludes unsupported reference classes. Recommendation precision,
recall, and coverage separate unsupported actions from missed supported ones. False-recommendation rate is conditional on
issued recommendations and undefined when none are issued. Eligibility and
evidence-state scores, VERIFY-to-EXCLUDE, MISSING-to-STALE and
MISSING-to-SUPPORTED rates identify different error types.

Risk--coverage curves group tied confidence values and use right-step area.
Let $g$ be the fraction of unresolved noncritical obligations. Initial \model{}
confidence is critical proof coverage multiplied by $1-0.05g$; it is not
calibrated risk. Proof rejections, citation traceability and inference efficiency accompany
decision quality.

\model{} is compared with Rule-only, Critic-and-guard, All-field compiler and \model{} w/o verifier using ten thousand paired entity-cluster
bootstrap resamples and pointwise 95\% intervals. Scenario-cluster resampling is a
sensitivity analysis. Candidate-level McNemar tests use Holm correction across
the twelve-comparison family; shared sources and queries limit their independent
pair interpretation. Effect sizes and cluster uncertainty remain primary.
Results separate base, compositional and stress partitions, with domain,
language, service-family, stress-kind and query-requirement slices. Positive
recall is unidentified where references provide no positive support.

\begin{table*}[t]\centering\small\setlength{\tabcolsep}{3pt}
\begin{tabular}{l*{9}{c}}\toprule
 & \multicolumn{3}{c}{Base} & \multicolumn{3}{c}{Compositional} & \multicolumn{3}{c}{Stress}\\Method & F1 & P/R & FRR & F1 & P/R & FRR & F1 & P/R & FRR\\\midrule
Rule-only & \textbf{0.915} & 1.00/1.00 & 0.000 & 0.816 & 1.00/1.00 & 0.000 & \textbf{0.927} & 1.00/1.00 & 0.000\\
Direct & 0.618 & 0.09/1.00 & 0.914 & 0.772 & 0.35/1.00 & 0.652 & 0.866 & 0.83/0.93 & 0.168\\
Flat RAG & 0.620 & 0.09/1.00 & 0.909 & 0.763 & 0.34/1.00 & 0.656 & 0.850 & 0.88/0.88 & 0.120\\
Critic-and-guard & 0.825 & 0.98/0.67 & 0.018 & 0.886 & 1.00/0.81 & 0.000 & 0.708 & 0.99/0.65 & 0.011\\
All-field compiler & 0.570 & --/0.00 & -- & 0.623 & --/0.00 & -- & 0.563 & --/0.00 & --\\
\model{} w/o verifier & 0.914 & 1.00/0.94 & 0.000 & \textbf{0.889} & 1.00/1.00 & 0.000 & 0.918 & 1.00/0.96 & 0.000\\
\model{} & 0.900 & 1.00/0.87 & 0.000 & 0.864 & 1.00/0.87 & 0.000 & 0.900 & 1.00/0.88 & 0.000\\
\bottomrule\end{tabular}
\caption{Held-out results with GPT-5-mini as the common language-model backbone. F1 is fixed-label decision macro-F1; P/R is recommendation precision/recall; FRR is false recommendation rate among issued recommendations. Bold marks the highest F1 in each partition. References are automated research-policy labels.}\label{tab:main}\end{table*}

\begin{table*}[t]\centering\small\setlength{\tabcolsep}{3pt}
\begin{tabular}{llccc}\toprule
Split & Baseline & $\Delta$ F1 & Pointwise 95\% cluster CI & Holm $p$\\\midrule
\multirow{4}{*}{Base} & Rule-only & -0.0148 & [-0.0329, 0.0020] & 0.7813\\
 & Critic-and-guard & 0.0755 & [0.0282, 0.1293] & $3.37\times10^{-22}$\\
 & All-field compiler & 0.3303 & [0.3059, 0.3562] & $4.74\times10^{-12}$\\
 & \model{} w/o verifier & -0.0133 & [-0.0263, -0.0039] & 0.1652\\
\addlinespace
\multirow{4}{*}{Comp.} & Rule-only & 0.0485 & [-0.0028, 0.0931] & 0.7813\\
 & Critic-and-guard & -0.0214 & [-0.1084, 0.0469] & 0.8746\\
 & All-field compiler & 0.2414 & [0.1629, 0.2911] & 0.1652\\
 & \model{} w/o verifier & -0.0246 & [-0.0531, -0.0053] & 0.5000\\
\addlinespace
\multirow{4}{*}{Stress} & Rule-only & -0.0271 & [-0.0525, -0.0003] & 0.0012\\
 & Critic-and-guard & 0.1916 & [0.1000, 0.2866] & $8.59\times10^{-15}$\\
 & All-field compiler & 0.3371 & [0.2688, 0.4126] & $8.67\times10^{-27}$\\
 & \model{} w/o verifier & -0.0189 & [-0.0300, -0.0093] & 0.0068\\
\bottomrule\end{tabular}
\caption{\model{} minus each baseline. Confidence intervals (CIs) use 10,000 paired entity-cluster resamples. Holm-adjusted candidate-pair McNemar tests form one family of twelve; these tests are diagnostic under repeated entities and queries. Supplementary statistics report scenario-cluster sensitivity and undefined denominators.}\label{tab:statistics}\end{table*}

\section{Results}
\label{sec:results}
\subsection{Do typed obligations recover useful recall?}
Table~\ref{tab:main} compares all seven methods on the same held-out candidates and source evidence pool.
The all-field compiler issues no recommendations, making its precision
and false-recommendation rate undefined. Full \model{} achieves base and
stress recall of \BlindVerifiedBaseRecRecall{} and
\BlindVerifiedStressRecRecall{}, respectively, with no unsupported
recommendations under the policy references. It meets all declared blind
effectiveness and integrity floors, listed in the supplement. This comparison
establishes a system-level recovery from over-abstention; the development
ablations below examine individual components.

Recovery does not imply overall superiority. \model{} macro-F1 is
\BlindVerifiedBaseFone{}, \BlindVerifiedCompFone{}, and
\BlindVerifiedStressFone{} on base, compositional, and stress cases; the
rule-only baseline obtains \BlindRuleBaseFone{}, \BlindRuleCompFone{}, and
\BlindRuleStressFone{}. The paired entity-cluster interval for \model{} minus Rule-only
excludes zero on stress in favor of Rule-only, but includes zero on base and
compositional cases. \model{} also has lower macro-F1 than unverified \model{}
in every partition, with all three pointwise intervals excluding zero.
Table~\ref{tab:statistics} reports the full comparison family and
Holm-adjusted McNemar diagnostics. Against the All-field compiler, the
macro-F1 gains range from 0.2414 to 0.3371, with all three cluster intervals
above zero. In contrast, verification changes F1 by -0.0133, -0.0246, and
-0.0189 relative to the unverified variant. The pointwise cluster intervals
and adjusted McNemar tests answer different questions: the former summarize
macro-F1 differences under entity-level resampling, whereas the latter test
candidate-level correctness across the comparison family. Their conclusions
need not coincide, so we interpret effect sizes with the corresponding
cluster intervals.

\begin{table*}[t]\centering\small\setlength{\tabcolsep}{3pt}
\begin{tabular}{lrrr}\toprule
Observed pipeline event & Base & Comp. & Stress\\\midrule
Initial \model{} w/o verifier recommendations & 79 & 31 & 128\\
Final \model{} recommendations & 73 & 27 & 117\\
Supported recommendations withdrawn & 6 & 4 & 11\\
Unsupported recommendations withdrawn & 0 & 0 & 0\\
\model{} w/o verifier decision errors & 233 & 18 & 16\\
\model{} decision errors & 239 & 22 & 27\\
Verifier calls & 79 & 31 & 128\\
Adjudicator calls & 7 & 5 & 19\\
Adjudication restorations & 1 & 1 & 8\\
Candidates with rejected query values & 0 & 5 & 0\\
Candidates with rejected proof links & 4287 & 359 & 453\\
Accepted verifier blockers & 8 & 6 & 19\\
Rejected verifier blockers & 1 & 0 & 2\\
\bottomrule\end{tabular}
\caption{Mechanical blind pipeline diagnostics. Correctness is measured against automated policy references. Rejected proof links and query values are logged events, not independently annotated semantic errors. Verifier/adjudicator counts exclude repeated attempts caused by connection failures.}\label{tab:diagnostics}\end{table*}

\subsection{Which components matter, and how stable are the results?}
Table~\ref{tab:ablation} reports the development ablations. Making every
information gap blocking lowers macro-F1 across all three partitions.
Removing the verifier increases macro-F1 in each partition, consistent
with the blind comparison of \model{} with and without verification. Scope, time, and semantic-check ablations
have mixed effects; the table does not support a claim that every added
check improves average performance.
\begin{table}[tbp]\centering\small\setlength{\tabcolsep}{3pt}
\begin{tabular}{lrrr}\toprule
Method & Base & Comp. & Stress\\\midrule
\model{} & 0.898 & 0.622 & 0.910\\
w/o criticality & 0.918 & 0.634 & 0.901\\
w/o scope & 0.911 & 0.609 & 0.919\\
w/o time & 0.905 & 0.493 & 0.893\\
w/o proof checks & 0.914 & 0.569 & 0.928\\
w/o verifier & 0.924 & 0.634 & 0.936\\
All gaps blocking & 0.589 & 0.321 & 0.556\\
\bottomrule\end{tabular}
\caption{Development ablations (decision macro-F1). Each ``w/o'' row removes the named component; ``All gaps blocking'' treats every information gap as blocking. The verifier ablation equals \model{} w/o verifier. Compiler ablations share query extractions and link proposals.}\label{tab:ablation}\end{table}

The development comparison selected Rule-only as the strongest qualified
baseline: its unweighted three-partition mean macro-F1 was
\DevRuleMeanFone{}, compared with \DevUnverifiedMeanFone{} for \model{} w/o verifier and
\DevVerifiedMeanFone{} for \model{}. The development compositional subset has no
\exclude{} references, so its fixed-label macro-F1 ceiling is $2/3$.
That ceiling does not apply to the blind compositional partition.

Across three independent runs, \model{} base recall averaged
\StabilityMeanBaseRecall{} (standard deviation \StabilitySDBaseRecall{}),
with a minimum of \StabilityWorstBaseRecall{}. This variation indicates that
recall recovery is not consistently robust across runs, despite the favorable
result in the primary test evaluation.

\subsection{How informative is confidence?}
\begin{figure*}[t]\centering\includegraphics[width=\textwidth]{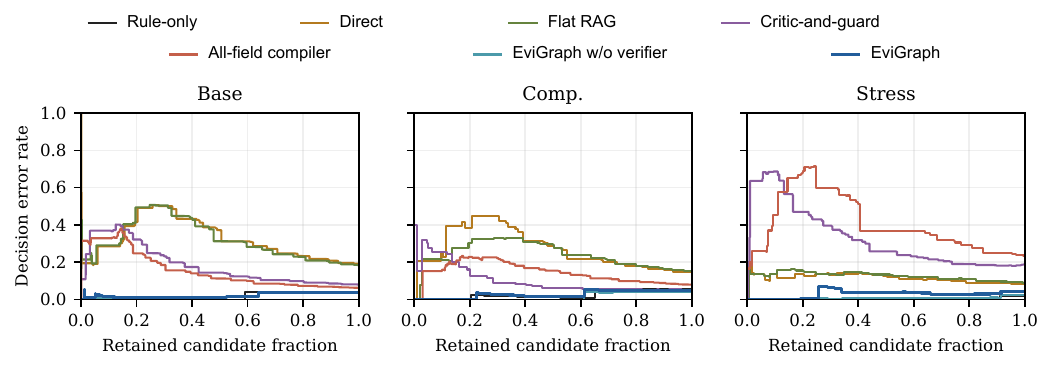}
\caption{Blind confidence-ranked risk--coverage curves. The ordinate is overall decision error on retained candidates, not false recommendation rate. Tied confidence values are grouped.}\label{fig:risk}\end{figure*}

Figure~\ref{fig:risk} shows overall decision error as confidence-ranked
coverage increases. Low error among issued recommendations does not ensure
low error among all retained candidates. In particular, the All-field compiler can make eligibility
or verification errors despite issuing no recommendations. Confidence rankings
also differ by partition, so these curves do not establish calibrated risk.

\section{Discussion}
\begin{table*}[t]\centering\small\setlength{\tabcolsep}{3pt}
\begin{tabular}{lcccccc}\toprule
Backbone & Direct & Flat RAG & \model{} & F1 & Rec. P/R & FRR\\\midrule
GPT-5 mini & $\checkmark$ & $\checkmark$ & $\checkmark$ & 0.900 & 1.00/0.87 & 0.000\\
Claude Opus 4.5 & $\checkmark$ & $\checkmark$ & $\checkmark$ & 0.944 & 1.00/0.98 & 0.000\\
Claude Sonnet 4.5 & $\checkmark$ & $\checkmark$ & $\checkmark$ & 0.902 & 1.00/0.80 & 0.000\\
\bottomrule\end{tabular}
\caption{Base-partition \model{} quality for backbones with complete results. GPT-5-mini is the primary evaluation; Claude results are exploratory.}\label{tab:models}
\end{table*}

\paragraph{Verification and supported recommendations.}
In the primary GPT-5-mini evaluation, \model{} w/o verifier issues \BlindInitialRecommendations{}
blind recommendations; \model{} retains \BlindFinalRecommendations{}.
Table~\ref{tab:diagnostics} separates these transitions by partition.
Verification withdraws \BlindSupportedWithdrawals{} policy-supported and
\BlindUnsupportedWithdrawals{} unsupported recommendations after
\BlindAdjudicatorCalls{} adjudications restore
\BlindAdjudicationRestorations{}. The verifier therefore reduces coverage
without an observed reduction in unsupported actions. A citation-valid concern
alone does not establish that withdrawal improves accuracy; decision transitions
must be evaluated separately from the number of criticisms produced.
Prior work distinguishes tool-supported correction \citep{gou2024critic}
from unreliable intrinsic revision \citep{huang2024selfcorrect}; our result
shows that access to evidence alone does not establish a decision benefit.

\paragraph{Proof checks and semantic correctness.}
Scope, time, and source attribution can be checked mechanically, but a quoted
person fact can still be misinterpreted. Table~\ref{tab:diagnostics} shows that
decision errors increase from 233 to 239 on base, 18 to 22 on compositional,
and 16 to 27 on stress cases, matching the supported recommendations withdrawn.
A candidate can retain a valid witness after an invalid link is discarded,
so rejected-link counts do not measure final decision errors or establish
independently annotated semantic causes.

\paragraph{Support and distribution boundaries.}
Positive recommendation references occur only in kindergarten vacancy navigation.
Other care domains lack resolved provider-specific requirements and cannot
identify positive-class recall (Table~\ref{tab:data}). Temporal contrasts
provide positive controls under fixed source evidence. This design and
support-enriched development sampling do not estimate natural
request or website-error prevalence.
Rule-only also benefits from programmatic query structure: its development scope
ablation reaches F1 \DevRuleNoScopeBaseFone{}, while removing time checks
raises compositional false-recommendation rate to \DevRuleNoTimeCompRisk{}.

\paragraph{Cross-backbone results.}
In the exploratory comparison (Table~\ref{tab:models}), Opus \model{} reaches
base F1 \ExtOpusVerifiedBaseFone{} and Sonnet \ExtSonnetVerifiedBaseFone{}, both
with zero observed false recommendations. Both exceed their own direct and
flat-RAG F1 throughout. Verification leaves Opus decision metrics unchanged
and lowers Sonnet F1, providing no observed gain across completed
configurations. The appendix reports partition results, paired intervals,
and repetition stability.
Each verified variant starts from its own backbone's initial proofs, enabling
a direct comparison of verification within that configuration. Cross-backbone
differences also reflect provider interfaces and reasoning settings, which are
not controlled here. The recurring coverage cost therefore supports separate
evaluation of verification, while its generality across services remains unresolved.

\paragraph{What verification can correct.}
Verification acts only on initial recommendations, so its potential benefit
is confined to detecting unsupported positive decisions. It cannot recover
supported candidates already assigned \verify{} or \exclude{}.
When initial recommendations are all policy-supported, there are no false
positives to remove, and accepted downgrades can only reduce recall.
This asymmetry motivates evaluating a verifier conditional on the errors
available for it to correct, alongside aggregate decision accuracy.

\section{Conclusion and Future Work}
Typed obligations recovered recall relative to an all-field compiler.
Verification reduced macro-F1 in all GPT-5-mini partitions and yielded no
gain with Claude, supporting separate assessment of proof construction
and verification.

A supported recommendation requires evidence for the requested service,
scope, and time; unresolved paperwork or preferences can remain follow-up
items. Valid citations cannot rescue misinterpreted requests, and withdrawing
supported options need not prevent unsupported actions. Evaluations should
report whether decision changes were justified and how much supported coverage
was retained.

Future work should use independently authored requests and broader
provider-specific evidence, with human assessment of request interpretation,
evidence applicability, and navigation utility. Targeted verification of
uncertain critical conditions and users' understanding of unresolved evidence
versus ineligibility warrant study before claiming improved service access.

\label{contentend}
\clearpage
\section*{Limitations}
\label{sec:limitations}
Requests are synthetic or programmatically composed; references are automated
policy judgments. Neither human query understanding nor actual applications,
admissions, reservations or service outcomes are evaluated. English and
Traditional Chinese renderings have no independent bilingual annotation, so
policy or rendering errors can affect both task definition and measured agreement.

Evidence is historical, and freshness windows are research assumptions.
A supported vacancy fact is not a current service guarantee. Sparse entities
and unresolved provider evidence limit distributional coverage and recall
identifiability. Entity relations may omit affiliations; structural-template
isolation does not prove complete semantic separation.

Candidate selection and temporal contrasts shape the evaluation distribution
and limit comparisons with differently constructed benchmarks. The backbone comparisons also differ in reasoning
and serving configurations, so their results cannot isolate the effect of
model choice. Findings are limited to the evaluated methods, domains, and
partitions.

\section*{Ethical Considerations}
The system is a research instrument for navigating public evidence. It is not
an automated eligibility authority and should not be used to deny access to
care. Verification means that evidence is insufficient for the stated request,
not that the person is unsuitable. Requests are synthetic; public provider
details and their source attribution are retained without implying endorsement
by the originating agencies. Source documents can have different reuse terms,
so redistribution distinguishes permitted derived annotations from source
materials that require source-specific handling. No claim of social benefit or
improved real-world access is made without a corresponding study.

\section*{Declaration on Generative AI}
A generative AI assistant was used to polish the language and presentation of
the manuscript.
\label{mainbodyend}
\bibliography{references}
\appendix
\Needspace{6\baselineskip}
\section{Evaluation Integrity}
All methods are evaluated on the same candidates and underlying source evidence
pool; retrieval and representation differ by method. In the primary evaluation,
reference labels are withheld until predictions have been finalized. Checks
confirm split separation, evidence attribution, and candidate coverage.
The exploratory cross-backbone analysis and its completeness criteria are
described in Appendix~\ref{sec:backboneextension}.

The supplementary materials retain model responses, failed attempts, and the
information needed to recompute the reported measures. Unavailable responses
are not replaced by invented predictions. Recomputing a score from retained
predictions is distinct from rerunning a language model, whose output may vary.

\Needspace{6\baselineskip}
\section{Additional Decision Examples}
\paragraph{Current versus future availability.}
A current vacancy leaves a future-date request unresolved. Removing the date
requirement can close that gap only if every other critical obligation is
supported. Capacity and retrieval recency cannot replace date-specific evidence.

\paragraph{Procedure versus hard eligibility.}
A missing medical form can require procedural follow-up; a demonstrated age
violation can establish ineligibility. Missing information does not prove the
latter, including when a controlled perturbation removes the applicable rule.

\paragraph{Requested versus attended grade.}
Requesting a kindergarten grade does not establish attendance at that grade.
A policy that permits attendance as an alternative to an age interval requires
an explicit attendance fact. Missing attendance is not proved nonattendance.

\Needspace{6\baselineskip}
\section{Interpretation of Reproducibility}
The evidence collection, reference rules, candidate selection, and proof
conditions define the evaluated setting. Reproducing the reported decisions
makes these choices inspectable; it does not validate them against human
judgments. Negative results remain part of the evaluation, and the conditions
for redistributing source materials are documented with the supplement.

\Needspace{6\baselineskip}
\section{Benchmark Construction}
Table~\ref{tab:validationfull} distinguishes validation from blind support;
recommendation counts describe policy references, not model predictions.
The benchmark contains \BaseCount{} Base, \CompositionalCount{} Compositional,
and \StressCount{} Stress scenarios, totaling \ScenarioCount{} scenarios and
\ScoredCount{} candidate decisions. Candidate lists are entity-disjoint across
splits. Reference labels are generated by explicit research-policy rules
rather than by human judgments.
\begin{table}[b]
\centering\small\setlength{\tabcolsep}{3pt}
\begin{tabular}{llrrr}
\toprule
Split & Partition & Queries & Scored & Rec. \\
\midrule
\multirow{3}{*}{Val.} & Base & 2,444 & 12,220 & 67 \\
 & Compositional & 133 & 665 & 26 \\
 & Stress & 691 & 691 & 86 \\
\addlinespace
\multirow{3}{*}{Blind} & Base & 1,228 & 6,140 & 84 \\
 & Compositional & 92 & 460 & 31 \\
 & Stress & 606 & 606 & 133 \\
\bottomrule\end{tabular}
\caption{Evaluation support. Rec. counts automated-policy recommendation candidates, not model successes. Compositional includes multi-requirement requests and current/future contrasts.}\label{tab:validationfull}
\end{table}

A structural query relation combines domain, service family, and the set of
constraint-field names. Replacing ages or dates, translating a query, or adding
a neutral introductory sentence does not create a new relation. Entity clusters,
service identifiers, and structural query relations are split-disjoint. Stress
variants retain their parent relation. The source-feasibility assignment depends
on visible source attributes, before references or model scores are computed.

The Compositional partition comprises \InheritedChallengeCount{} requests
combining multiple requirements and \AddedChallengeCount{} explicitly marked
temporal contrasts. The former supply only \InheritedPositiveCount{} positive
candidate references, motivating temporal contrasts as positive controls.
Each domain contributes fifty contrast scenarios organized into twenty-five
current/future pairs, without changing the source facts. Positive controls
come from kindergarten navigation; provider-specific care capabilities remain
unresolved in other domains. This support-oriented construction does not
estimate the prevalence of real care requests.

No base recommendation examples occur in the fit or calibration split. These
splits therefore cannot establish positive-class risk calibration. The method
uses a declared proof-coverage confidence score and makes no calibrated-risk
claim. All positive recommendation references in the available evidence are in
kindergarten navigation; other domain slices cannot measure positive recall.

\Needspace{6\baselineskip}
\section{Query and Proof Interfaces}
Each query requirement specifies its meaning, decision role, necessity,
value, scope, source, and an exact quote from the request.
One scenario-level extraction is reused for every candidate. The model cannot
change person facts or requested dates according to which provider is examined.

The checker verifies that each interpreted value has the required type. Service family, requested grade, attendance grade, requested date,
time interval and session must normalize to strings; ages and quantitative
limits normalize to numbers; explicit binary person facts normalize to booleans.
Malformed values are rejected and unresolved requirements are retained. They
are not repaired from privileged construction information or expected decisions. Requested kindergarten grade never proves actual attendance.

The local linker sees the shared query requirements, the candidate's visible
records, and a catalog of obligation identifiers, fields and types. The catalog
contains no reference status or expected answer. The linker proposes PASS,
FAIL, or UNKNOWN and candidate-local rule and claim identifiers. A failed
hypothesis is not an instruction to redraw an answer: missing and unknown links
are handled by deterministic proof checking. Evidence must be uniquely attributable to the candidate, and all required
candidates must be represented.

Every evidence record has a provenance path. Service/grade/session/date scope
and effective update time remain distinct from retrieval time. Capacity and a
service registry listing cannot close a vacancy obligation. A current vacancy
cannot close a future-date availability obligation. The checker retains one
representative of equivalent supporting observations and each distinct required
rule condition. This is minimality within the declared witness representation,
not a globally shortest graph proof.

Validated positive witnesses use SUPPORTS edges and hard failures use
CONTRADICTS edges. Invalid identifiers cannot create new graph nodes. A proof is
relative to the normalized query and published research policy; its existence
is not evidence that a provider will actually admit or reserve a place.

\Needspace{6\baselineskip}
\section{Method Configurations}
Rule-only uses a fixed bilingual regular-expression extractor and deterministic rule
comparisons. It shares the published decision algebra with \model{} but makes
no language-model calls. Template-generated queries favor its fixed extractor,
which limits generalization to independently authored requests.

Direct receives flat evidence and the typed navigation policy. Flat RAG retrieves at most
twelve claim records per candidate by BM25, with $k_1=1.2$ and $b=0.75$, and
retains all published rules. Lexical scores use attribute, value and scope;
source-record names do not contribute. English words and individual Chinese, Japanese, and Korean (CJK)
characters form the fixed tokenization. Flat RAG is a flat retrieval baseline, not an
embedding-model comparison.

Critic-and-guard combines a compact evidence representation, recommendation
critic, and future-availability guard. The All-field compiler maps structured
assessments to decisions with unresolved assessment fields acting as blockers.
Its semantic checks retain scope and time constraints. Provider-background
assertions remain visible but cannot establish hard eligibility failures.

\model{} w/o verifier performs shared extraction, local linking and typed compilation. \model{} adds
verification only for initial recommendations. A verifier blocker must name an
existing critical obligation, an edge in its validated path, visible evidence,
and a specific applicability or temporal error. New paperwork requirements
and noncritical gaps are invalid blockers. At most one adjudication may restore
an initially recommended candidate by addressing every accepted blocker with
existing proof-edge identifiers. Initial VERIFY and EXCLUDE candidates cannot
be upgraded by this stage.

The criticality ablation treats procedural obligations as critical; the scope ablation removes semantic scope checks;
the time ablation removes temporal checks; the proof-check ablation trusts semantic PASS/FAIL hypotheses while keeping
the common candidate-local identifier contract; the verifier ablation equals \model{} w/o verifier; the all-gaps-blocking ablation makes all gaps
blocking. Query extractions and local-link responses are shared across ablations.
Verifier workloads are recorded separately.

\Needspace{6\baselineskip}
\section{Execution and Reproducibility}
\begin{table*}[t]\centering\small\setlength{\tabcolsep}{3pt}
\begin{tabular}{lrrrrr}\toprule
Method & Calls & In/Out (M tokens) & p95 (s) & Invalid & Retries\\\midrule
Rule-only & 0 & 0.00/0.00 & -- & 0 & 0\\
Direct & 1941 & 49.03/5.38 & 54.9 & 0 & 15\\
Flat RAG & 1939 & 32.35/5.38 & 55.8 & 1 & 13\\
Critic-and-guard & 2337 & 36.52/6.89 & 114.2 & 0 & 15\\
All-field compiler & 1972 & 71.26/19.85 & 351.6 & 27 & 46\\
\model{} w/o verifier & 3913 & 40.47/14.48 & 109.0 & 29 & 61\\
\model{} & 4184 & 47.82/15.08 & 107.4 & 29 & 63\\
\bottomrule\end{tabular}
\caption{GPT-5-mini inference resource use. Latency is per observed request, not end-to-end user latency. All language-model methods use Flex; separate execution windows limit latency comparisons. \model{} w/o verifier and \model{} share extraction and linking. Invalid denotes responses that failed validation.}\label{tab:efficiency}\end{table*}

Table~\ref{tab:efficiency} reports resource use for the primary evaluation.
GPT-5-mini uses snapshot \texttt{gpt-5-mini-2025-08-07} with medium reasoning
effort. Development repetitions use seeds 20260908, 20260909, and 20260910.
Development selection and test evaluation use different service tiers, so
latency is descriptive rather than a controlled deployment comparison.
Model configurations and raw responses are retained in the research
archive. Reference labels are excluded from model inputs, and only complete,
validated predictions are scored.

\paragraph{Response validation and retries.}
Before test-reference access, the execution policy was amended to extend
bounded retries for connection failures and mechanically invalid scenario or
candidate identifiers in the All-field compiler and \model{}. Eligible identity
failures permit nine attempts in total; other invalid responses remain capped
at three. Semantic proof errors do not qualify for the identity exception.
This exception changed the prespecified retry budget, not the prompts, visible
evidence, model settings, or validators. Each retransmission preserves the
serialized request; valid responses are reused, and identifiers or semantic
predictions are never manually repaired. Eligibility depends on mechanical
validation rather than predicted correctness or reference labels. Retries seek
a complete response satisfying the method's output contract; they do not
select among valid predictions by quality. Table~\ref{tab:efficiency} reports
aggregate retries and resource use, not per-request allowances. Unequal retry
workloads remain a condition of the comparison, and all attempts are archived.

\Needspace{6\baselineskip}
\section{Development Results}
\begin{table}[tbp]\centering\small\setlength{\tabcolsep}{3pt}
\begin{tabular}{lrrr}\toprule
Method & Base & Comp. & Stress\\\midrule
Rule-only & 0.952 & 0.667 & 0.978\\
Direct & 0.801 & 0.434 & 0.861\\
Flat RAG & 0.815 & 0.416 & 0.896\\
Critic-and-guard & 0.801 & 0.581 & 0.647\\
All-field compiler & 0.569 & 0.321 & 0.580\\
\model{} w/o verifier & 0.924 & 0.634 & 0.936\\
\model{} & 0.898 & 0.622 & 0.910\\
\bottomrule\end{tabular}
\caption{Development decision macro-F1 on the fixed support-enriched subset (100 base, 50 compositional and 100 stress queries). Results use the standard serving tier and seed 20260908 for method selection, separately from held-out testing.}\label{tab:development}\end{table}

Table~\ref{tab:development} shows that the rule-only method leads the
development comparison, while Table~\ref{tab:m0ablation} tests which of its checks
contribute to that performance. The fixed subset is enriched for policy-reference recommendation support and
does not estimate the full split's natural frequency. The strongest qualified
baseline is Rule-only. \model{} passes the declared development floors but has lower decision
macro-F1 than Rule-only and \model{} w/o verifier on all three partitions. Passing a minimum floor does
not support a claim of superiority.
The development compositional subset has no EXCLUDE references. With the fixed
three-class convention, its best attainable macro-F1 is therefore $2/3$;
present-reference macro-F1 and per-class support are retained in the metrics.
This subset-specific ceiling does not apply to the full blind compositional
partition, which contains EXCLUDE cases.
\begin{table*}[t]\centering\small\setlength{\tabcolsep}{3pt}
\begin{tabular}{lrrrrr}\toprule
Rule-only variant & Base F1 & Comp. F1 & Stress F1 & Comp. FRR & Stress FRR\\\midrule
Full & 0.952 & 0.667 & 0.978 & 0.000 & 0.000\\
w/o criticality & 0.952 & 0.667 & 0.978 & 0.000 & 0.000\\
w/o scope & 1.000 & 0.667 & 0.978 & 0.000 & 0.000\\
w/o time & 0.952 & 0.515 & 0.938 & 0.571 & 0.143\\
w/o proof checks & 0.952 & 0.667 & 0.978 & 0.000 & 0.000\\
w/o verifier & 0.952 & 0.667 & 0.978 & 0.000 & 0.000\\
All gaps blocking & 0.952 & 0.667 & 0.978 & 0.000 & 0.000\\
\bottomrule\end{tabular}
\caption{Strongest-baseline development ablations, all deterministic and API-free. The scope and time ablations disable checks in both obligation preparation and compilation. The verifier ablation is an identity because Rule-only has no verifier. The proof-check ablation tests validation of Rule-only's deterministic proposals. Rule-only extracts fewer noncritical query fields than the LLM extractor, limiting comparisons of criticality and all-gaps-blocking effects across systems.}\label{tab:m0ablation}\end{table*}

Rule-only ablations make no API calls. Its narrower extractor does not represent all
noncritical fields produced by the large language model (LLM) extractor; the criticality and all-gaps-blocking ablations therefore leave its
decisions unchanged. The verifier ablation is also an identity because Rule-only has no verifier. Removing
time checks increases its compositional false-recommendation rate to \DevRuleNoTimeCompRisk{}
and stress rate to \DevRuleNoTimeStressRisk{}. Removing scope checks improves its base macro-F1 to
\DevRuleNoScopeBaseFone{} in this development subset, illustrating that a check's
contribution depends on the evaluated query distribution.

\Needspace{6\baselineskip}
\section{Evaluation Measures and Statistical Interpretation}
Fixed-label decision macro-F1 averages RECOMMEND, VERIFY and EXCLUDE, assigning
zero F1 to an absent class. Present-reference macro-F1 excludes unsupported
reference classes. Recommendation precision and false-recommendation rate are
undefined with no predicted recommendations; recall is undefined with no
reference recommendations. Undefined denominators cannot satisfy a safety or
recall gate by convention.

Confidence is critical-obligation proof coverage multiplied by
$1-0.05g$, where $g$ is the fraction of unresolved procedural, preference and
informational obligations. This penalty can lower confidence without changing
a recommendation. Risk--coverage curves group all equal-confidence cases and
use the resulting right-step area. They do not establish a calibrated risk
bound.

The primary paired bootstrap samples entity clusters with replacement and
retains all candidate rows in each sampled cluster. Both compared methods share
the same resampling weights. Ten thousand resamples yield pointwise percentile 95\%
intervals for measure differences. Scenario-cluster resampling is a separate
sensitivity analysis. Undefined bootstrap denominators remain undefined, and
the number of defined replicates is reported.

\model{} is compared with Rule-only, Critic-and-guard, All-field compiler and \model{} w/o verifier across three partitions. Candidate-pair
correctness McNemar tests form one family of twelve Holm-corrected comparisons.
These candidate-level tests are diagnostic because repeated entities and queries
may violate independent-pair assumptions; cluster intervals and effect sizes
carry the primary interpretation. Three-seed summaries report the mean,
population standard deviation and worst observed seed. Repeated deterministic
Rule-only executions are not independent stochastic replicates.

\Needspace{6\baselineskip}
\section{Blind Results and Stability}
\begin{table*}[t]\centering\small\setlength{\tabcolsep}{3pt}
\begin{tabular}{llrrrr}\toprule
Split & Group & Queries & Candidates & R/V/E & S/C/T/M\\\midrule
\multirow{3}{*}{Base} & Child/family & 742 & 3710 & 84/3586/40 & 110/0/0/3600\\
 & Disability & 204 & 1020 & 0/737/283 & 0/0/0/1020\\
 & Elderly & 282 & 1410 & 0/1070/340 & 0/0/0/1410\\
\addlinespace
\multirow{3}{*}{Comp.} & Child/family & 53 & 265 & 31/224/10 & 40/0/0/225\\
 & Disability & 30 & 150 & 0/122/28 & 0/0/0/150\\
 & Elderly & 9 & 45 & 0/45/0 & 0/0/0/45\\
\addlinespace
\multirow{3}{*}{Stress} & Child/family & 466 & 466 & 133/331/2 & 178/122/55/111\\
 & Disability & 42 & 42 & 0/33/9 & 0/0/0/42\\
 & Elderly & 98 & 98 & 0/77/21 & 0/44/0/54\\
\midrule
\multirow{3}{*}{Base} & Bilingual & 846 & 4230 & 52/3725/453 & 70/0/0/4160\\
 & English & 192 & 960 & 21/807/132 & 26/0/0/934\\
 & Trad. Chinese & 190 & 950 & 11/861/78 & 14/0/0/936\\
\addlinespace
\multirow{3}{*}{Comp.} & Bilingual & 62 & 310 & 26/264/20 & 34/0/0/276\\
 & English & 13 & 65 & 3/59/3 & 3/0/0/62\\
 & Trad. Chinese & 17 & 85 & 2/68/15 & 3/0/0/82\\
\addlinespace
\multirow{3}{*}{Stress} & Bilingual & 411 & 411 & 80/316/15 & 113/120/34/144\\
 & English & 111 & 111 & 29/69/13 & 37/26/12/36\\
 & Trad. Chinese & 84 & 84 & 24/56/4 & 28/20/9/27\\
\bottomrule\end{tabular}
\caption{Blind benchmark support by domain and language. R/V/E denotes automated-policy RECOMMEND/VERIFY/EXCLUDE; S/C/T/M denotes SUPPORTED/CONFLICTING/STALE/MISSING. Domain and language are separate overlapping views and must not be summed together. Construction totals and the full validation split are reported in the main benchmark table.}\label{tab:benchmarksupport}\end{table*}

\begin{table*}[t]\centering\small\setlength{\tabcolsep}{3pt}
\begin{tabular}{llccccc}\toprule
Method & Split & F1 $\mu\pm\sigma$ & Worst F1 & Rec. R $\mu\pm\sigma$ & Worst R & Worst FRR\\\midrule
\multirow{3}{*}{Rule-only} & Base & 0.952$\pm$0.000 & 0.952 & 1.000$\pm$0.000 & 1.000 & 0.000\\
 & Comp. & 0.667$\pm$0.000 & 0.667 & 1.000$\pm$0.000 & 1.000 & 0.000\\
 & Stress & 0.978$\pm$0.000 & 0.978 & 1.000$\pm$0.000 & 1.000 & 0.000\\
\addlinespace
\multirow{3}{*}{\model{}} & Base & 0.899$\pm$0.024 & 0.879 & 0.833$\pm$0.115 & 0.688 & 0.000\\
 & Comp. & 0.645$\pm$0.016 & 0.622 & 0.889$\pm$0.079 & 0.778 & 0.000\\
 & Stress & 0.949$\pm$0.013 & 0.934 & 0.833$\pm$0.027 & 0.800 & 0.000\\
\bottomrule\end{tabular}
\caption{Development stability over three fixed seeds, 20260908/09/10. Worst denotes the minimum F1/recall or maximum FRR. \model{} uses three independent Flex runs. Rule-only is deterministic: identical repeats measure reproducibility and carry no independent stochastic interpretation. Standard-tier method-selection results are reported separately.}\label{tab:stability}\end{table*}

\begin{table*}[t]\centering\small\setlength{\tabcolsep}{3pt}
\begin{tabular}{llcc}\toprule
\model{} criterion & Declared floor & Observed & Met\\\midrule
Base FRR & $\leq .05$ & 0.000 & Yes\\
Stress FRR & $\leq .05$ & 0.000 & Yes\\
Base Rec. R & $\geq .72$ & 0.869 & Yes\\
Stress Rec. R & $\geq .80$ & 0.880 & Yes\\
Base V-to-E & $\leq .03$ & 0.000 & Yes\\
Stress V-to-E & $\leq .05$ & 0.000 & Yes\\
Stress Conflict R & $\geq .65$ & 1.000 & Yes\\
Stress Stale P & $\geq .60$ & 0.962 & Yes\\
\bottomrule\end{tabular}
\caption{Predeclared \model{} effectiveness floors on blind data. V-to-E is the fraction of policy-reference VERIFY candidates predicted EXCLUDE. Undefined denominators do not pass. Floor failures are reported as negative results and never trigger test-set tuning. Schema validity and candidate-local citation traceability are checked separately.}\label{tab:floors}\end{table*}

\begin{figure*}[t]\centering\includegraphics[width=\textwidth]{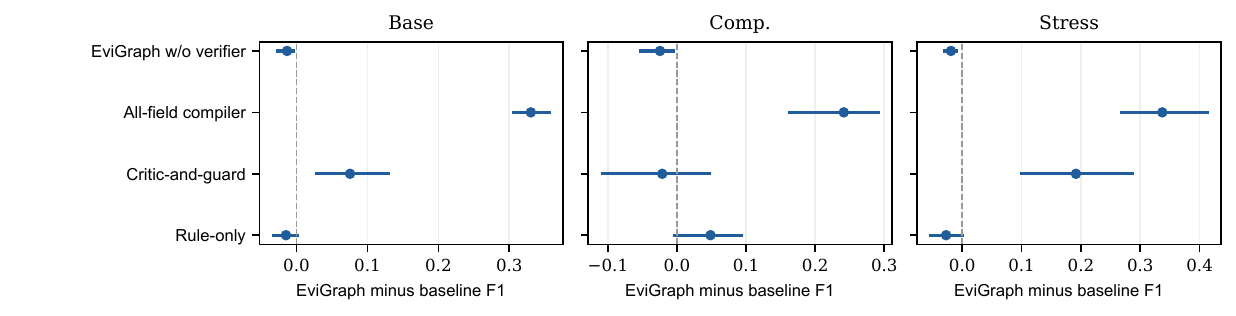}
\caption{Paired entity-cluster 95\% intervals for \model{} minus each baseline in blind decision macro-F1; 10,000 resamples. Zero marks equal performance.}\label{fig:effects}\end{figure*}

\begin{table*}[t]\centering\small\setlength{\tabcolsep}{3pt}
\begin{tabular}{llrrrrrrr}\toprule
Split & Method & $N$ & Rec. support & Present F1 & Elig. F1 & Evid. F1 & Cite & AURC\\\midrule
\multirow{7}{*}{Base} & Rule-only & 6140 & 84 & 0.915 & 0.837 & 0.497 & 1.000 & 0.017\\
 & Direct & 6140 & 84 & 0.618 & 0.714 & 0.209 & 0.990 & 0.320\\
 & Flat RAG & 6140 & 84 & 0.620 & 0.714 & 0.212 & 0.984 & 0.316\\
 & Critic-and-guard & 6140 & 84 & 0.825 & 0.655 & 0.245 & 1.000 & 0.181\\
 & All-field compiler & 6140 & 84 & 0.570 & 0.582 & 0.364 & 1.000 & 0.156\\
 & \model{} w/o verifier & 6140 & 84 & 0.914 & 0.871 & 0.458 & 1.000 & 0.020\\
 & \model{} & 6140 & 84 & 0.900 & 0.871 & 0.458 & 1.000 & 0.022\\
\addlinespace
\multirow{7}{*}{Comp.} & Rule-only & 460 & 31 & 0.816 & 0.786 & 0.500 & 1.000 & 0.027\\
 & Direct & 460 & 31 & 0.772 & 0.711 & 0.269 & 0.956 & 0.267\\
 & Flat RAG & 460 & 31 & 0.763 & 0.678 & 0.265 & 0.968 & 0.238\\
 & Critic-and-guard & 460 & 31 & 0.886 & 0.632 & 0.361 & 1.000 & 0.103\\
 & All-field compiler & 460 & 31 & 0.623 & 0.604 & 0.378 & 1.000 & 0.143\\
 & \model{} w/o verifier & 460 & 31 & 0.889 & 0.871 & 0.466 & 1.000 & 0.015\\
 & \model{} & 460 & 31 & 0.864 & 0.871 & 0.466 & 1.000 & 0.027\\
\addlinespace
\multirow{7}{*}{Stress} & Rule-only & 606 & 133 & 0.927 & 0.918 & 0.989 & 1.000 & 0.002\\
 & Direct & 606 & 133 & 0.866 & 0.609 & 0.772 & 0.994 & 0.116\\
 & Flat RAG & 606 & 133 & 0.850 & 0.642 & 0.781 & 0.994 & 0.126\\
 & Critic-and-guard & 606 & 133 & 0.708 & 0.557 & 0.709 & 1.000 & 0.327\\
 & All-field compiler & 606 & 133 & 0.563 & 0.424 & 0.792 & 1.000 & 0.401\\
 & \model{} w/o verifier & 606 & 133 & 0.918 & 0.917 & 0.822 & 1.000 & 0.007\\
 & \model{} & 606 & 133 & 0.900 & 0.917 & 0.822 & 1.000 & 0.029\\
\bottomrule\end{tabular}
\caption{Additional blind metrics. Cite is source traceability for each candidate, which does not establish semantic correctness. Present F1 excludes unsupported reference classes. AURC measures confidence-ordered overall decision error with tied scores grouped. Complete breakdowns by domain, language, service family, and stress type accompany the supplementary data.}\label{tab:extended}\end{table*}

Positive recommendation references occur only in child/family services
(Table~\ref{tab:benchmarksupport}).
Table~\ref{tab:extended} distinguishes decision agreement from eligibility,
evidence-state, and citation measures: perfect source traceability does not
ensure correct decisions.

\model{} meets the blind effectiveness floors (Table~\ref{tab:floors}), yet
verification withdraws \BlindSupportedWithdrawals{} supported recommendations
without removing unsupported ones (Table~\ref{tab:diagnostics}). All three
paired macro-F1 intervals favor the unverified variant
(Table~\ref{tab:statistics}; Figure~\ref{fig:effects}). This loss concerns useful
coverage, not citation validity; mechanical rejection counts alone cannot
identify semantic error causes.

Development base recall averages $0.833\pm0.115$ and falls to 0.688 in the
worst run, despite zero observed false recommendations
(Table~\ref{tab:stability}). Seed 20260909 therefore fails the base-recall floor;
averaging does not erase that failure. High precision can coexist with variable
supported coverage, motivating joint reporting of precision, recall, and
worst-run performance.

\Needspace{6\baselineskip}
\section{Held-Out Evaluation and Reproducibility}
Benchmark checks verify split separation, evidence attribution, candidate
coverage, and sufficient positive examples. These checks establish that an
evaluation can be performed; they do not establish a method's effectiveness.

In the primary evaluation, all methods must produce valid predictions for the
same candidates before scoring. Incorrect baseline citations are counted
without repairing predictions. \model{} must satisfy its source-traceability
requirements. Only after predictions are finalized can reference labels be
used to calculate scores. Test results cannot change the selected prompts,
evidence roles, decision thresholds, or experimental settings.

The supplementary materials distinguish public evidence, model inputs,
reference judgments, responses, and evaluation results. They support inspection
and recomputation of the reported findings while respecting source-specific
redistribution conditions. They do not guarantee that future model responses
or public-service information will remain unchanged.
\Needspace{6\baselineskip}
\section{Exploratory Cross-Backbone Evaluation}
\label{sec:backboneextension}
The cross-backbone evaluation uses the fixed 250-scenario development subset and all
1,926 test scenarios (7,206 candidate decisions). No test candidate subset is
selected for comparison. Primary GPT-5-mini outputs are reused without new
benchmark inference. The additional models are
\texttt{gpt-4o-mini-2024-07-18},
\texttt{claude-sonnet-4-5-20250929}, and
\texttt{claude-opus-4-5-20251101}.
The planned configurations are Direct, Flat RAG, and a shared \model{} execution
producing \model{} w/o verifier and \model{}; only complete units are scored.
Development has three repetitions labeled 20260908, 20260909, and 20260910;
test execution uses label 20260908. GPT-4o mini receives the API seed parameter.
Claude provides no corresponding seed parameter, so its labels identify
independent repetitions rather than controlled random seeds. All additional models
use temperature zero. Claude extended thinking is not enabled. The primary
GPT-5-mini reasoning configuration and Flex tier remain unchanged; these
experiments do not hold reasoning compute or service tier constant.

\begin{table*}[tp]\centering\small\setlength{\tabcolsep}{3pt}\begin{tabular}{llccc}\toprule Backbone & Method & Decision F1 & Rec. P/R & FRR\\\midrule
\multirow{4}{*}{GPT-5 mini} & Direct & 0.618 & 0.09/1.00 & 0.914\\
 & Flat RAG & 0.620 & 0.09/1.00 & 0.909\\
 & \model{} w/o verifier & 0.914 & 1.00/0.94 & 0.000\\
 & \model{} & 0.900 & 1.00/0.87 & 0.000\\
\addlinespace
\multirow{4}{*}{Claude Opus 4.5} & Direct & 0.647 & 0.45/0.15 & 0.552\\
 & Flat RAG & 0.682 & 0.40/0.32 & 0.597\\
 & \model{} w/o verifier & 0.944 & 1.00/0.98 & 0.000\\
 & \model{} & 0.944 & 1.00/0.98 & 0.000\\
\addlinespace
\multirow{4}{*}{Claude Sonnet 4.5} & Direct & 0.643 & 0.21/0.21 & 0.793\\
 & Flat RAG & 0.731 & 0.49/0.43 & 0.507\\
 & \model{} w/o verifier & 0.914 & 1.00/0.86 & 0.000\\
 & \model{} & 0.902 & 1.00/0.80 & 0.000\\
\addlinespace
Shared rule-only & Rule-only & 0.915 & 1.00/1.00 & 0.000\\\bottomrule\end{tabular}
\caption{Base: complete-partition cross-backbone results. Rule-only is shared and deterministic. References are automated policy judgments.}\label{tab:multi-base}\end{table*}

\begin{table*}[tp]\centering\small\setlength{\tabcolsep}{3pt}\begin{tabular}{llccc}\toprule Backbone & Method & Decision F1 & Rec. P/R & FRR\\\midrule
\multirow{4}{*}{GPT-5 mini} & Direct & 0.772 & 0.35/1.00 & 0.652\\
 & Flat RAG & 0.763 & 0.34/1.00 & 0.656\\
 & \model{} w/o verifier & 0.889 & 1.00/1.00 & 0.000\\
 & \model{} & 0.864 & 1.00/0.87 & 0.000\\
\addlinespace
GPT-4o mini & Flat RAG & 0.441 & 0.21/0.87 & 0.792\\
\addlinespace
\multirow{4}{*}{Claude Opus 4.5} & Direct & 0.674 & 0.67/0.26 & 0.333\\
 & Flat RAG & 0.817 & 0.79/0.74 & 0.207\\
 & \model{} w/o verifier & 0.974 & 1.00/1.00 & 0.000\\
 & \model{} & 0.974 & 1.00/1.00 & 0.000\\
\addlinespace
\multirow{4}{*}{Claude Sonnet 4.5} & Direct & 0.672 & 0.69/0.29 & 0.308\\
 & Flat RAG & 0.784 & 0.75/0.48 & 0.250\\
 & \model{} w/o verifier & 0.900 & 1.00/0.94 & 0.000\\
 & \model{} & 0.894 & 1.00/0.90 & 0.000\\
\addlinespace
Shared rule-only & Rule-only & 0.816 & 1.00/1.00 & 0.000\\\bottomrule\end{tabular}
\caption{Compositional: complete-partition cross-backbone results. Rule-only is shared and deterministic. References are automated policy judgments.}\label{tab:multi-compositional_challenge}\end{table*}

\begin{table*}[tp]\centering\small\setlength{\tabcolsep}{3pt}\begin{tabular}{llccc}\toprule Backbone & Method & Decision F1 & Rec. P/R & FRR\\\midrule
\multirow{4}{*}{GPT-5 mini} & Direct & 0.866 & 0.83/0.93 & 0.168\\
 & Flat RAG & 0.850 & 0.88/0.88 & 0.120\\
 & \model{} w/o verifier & 0.918 & 1.00/0.96 & 0.000\\
 & \model{} & 0.900 & 1.00/0.88 & 0.000\\
\addlinespace
\multirow{2}{*}{GPT-4o mini} & Direct & 0.440 & 0.43/0.89 & 0.574\\
 & Flat RAG & 0.469 & 0.44/0.97 & 0.563\\
\addlinespace
\multirow{4}{*}{Claude Opus 4.5} & Direct & 0.667 & 1.00/0.20 & 0.000\\
 & Flat RAG & 0.673 & 0.97/0.23 & 0.032\\
 & \model{} w/o verifier & 0.930 & 1.00/0.83 & 0.000\\
 & \model{} & 0.930 & 1.00/0.83 & 0.000\\
\addlinespace
\multirow{4}{*}{Claude Sonnet 4.5} & Direct & 0.616 & 0.88/0.17 & 0.120\\
 & Flat RAG & 0.719 & 0.78/0.32 & 0.222\\
 & \model{} w/o verifier & 0.931 & 1.00/0.95 & 0.000\\
 & \model{} & 0.908 & 1.00/0.85 & 0.000\\
\addlinespace
Shared rule-only & Rule-only & 0.927 & 1.00/1.00 & 0.000\\\bottomrule\end{tabular}
\caption{Controlled stress: complete-partition cross-backbone results. Rule-only is shared and deterministic. References are automated policy judgments.}\label{tab:multi-controlled_stress}\end{table*}

\paragraph{Evaluation scope and completeness.}
This exploratory analysis was designed after the primary GPT-5-mini test
results were known and is not an independent blind confirmation. Method
configurations and validation rules were fixed before additional inference;
statistical comparisons were specified before scoring the additional models.

Scoring requires complete model--method--repetition--partition units under
criteria specified before reference access. Each unit must cover the full
candidate set and pass unchanged validation rules; missing outputs are not
imputed. The unverified variant requires complete extraction and linking;
full \model{} also requires verification and adjudication. Predictions are
finalized before scoring, with no subsequent benchmark generation.

\begin{table*}[tp]\centering\small\setlength{\tabcolsep}{3pt}\begin{tabular}{p{0.53\textwidth}ccc}\toprule Comparison & $\Delta$F1 & Entity 95\% CI & Holm $p$\\\midrule
Claude Opus 4.5 \model{} vs. Direct & 0.2967 & [0.2509, 0.3486] & 2.3e-39\\
Claude Opus 4.5 \model{} vs. Flat RAG & 0.2623 & [0.2094, 0.3227] & 2.54e-49\\
Claude Opus 4.5 \model{} vs. \model{} w/o verifier & 0.0000 & [0.0000, 0.0000] & 1\\
Claude Opus 4.5 \model{} vs. GPT-5 mini \model{} & 0.0437 & [0.0271, 0.0621] & 1.87e-18\\
Claude Opus 4.5 \model{} vs. Rule-only & 0.0289 & [0.0191, 0.0390] & 1.22e-21\\
Claude Sonnet 4.5 \model{} vs. Direct & 0.2595 & [0.2138, 0.3098] & 1.17e-35\\
Claude Sonnet 4.5 \model{} vs. Flat RAG & 0.1714 & [0.1208, 0.2269] & 1.43e-23\\
Claude Sonnet 4.5 \model{} vs. \model{} w/o verifier & -0.0120 & [-0.0235, -0.0027] & 1\\
Claude Sonnet 4.5 \model{} vs. GPT-5 mini \model{} & 0.0021 & [-0.0227, 0.0259] & 0.00017\\
Claude Sonnet 4.5 \model{} vs. Rule-only & -0.0127 & [-0.0347, 0.0062] & 3.85e-06\\
GPT-5 mini \model{} vs. Direct & 0.2822 & [0.2470, 0.3242] & 3.32e-152\\
GPT-5 mini \model{} vs. Flat RAG & 0.2807 & [0.2442, 0.3235] & 5.5e-145\\
GPT-5 mini \model{} vs. \model{} w/o verifier & -0.0133 & [-0.0263, -0.0039] & 0.875\\
GPT-5 mini \model{} vs. Rule-only & -0.0148 & [-0.0329, 0.0020] & 1\\\bottomrule\end{tabular}
\caption{Base: paired exploratory cross-backbone comparisons. Intervals use 10,000 entity-cluster bootstrap replicates; McNemar $p$ values are diagnostic and Holm-corrected within the 57-comparison family. Only the 42 completed comparisons are reported; correction retains the full family of 57 prespecified comparisons. Scenario-cluster sensitivity results are archived.}\label{tab:multi-stat-base}\end{table*}

\begin{table*}[tp]\centering\small\setlength{\tabcolsep}{3pt}\begin{tabular}{p{0.53\textwidth}ccc}\toprule Comparison & $\Delta$F1 & Entity 95\% CI & Holm $p$\\\midrule
Claude Opus 4.5 \model{} vs. Direct & 0.3000 & [0.2167, 0.3914] & 2.54e-10\\
Claude Opus 4.5 \model{} vs. Flat RAG & 0.1576 & [0.0975, 0.2296] & 1.94e-06\\
Claude Opus 4.5 \model{} vs. \model{} w/o verifier & 0.0000 & [0.0000, 0.0000] & 1\\
Claude Opus 4.5 \model{} vs. GPT-5 mini \model{} & 0.1100 & [0.0571, 0.1910] & 0.000565\\
Claude Opus 4.5 \model{} vs. Rule-only & 0.1585 & [0.1073, 0.2407] & 3.81e-05\\
Claude Sonnet 4.5 \model{} vs. Direct & 0.2215 & [0.1220, 0.3116] & 3.06e-05\\
Claude Sonnet 4.5 \model{} vs. Flat RAG & 0.1094 & [0.0060, 0.1914] & 0.226\\
Claude Sonnet 4.5 \model{} vs. \model{} w/o verifier & -0.0062 & [-0.0224, 0.0000] & 1\\
Claude Sonnet 4.5 \model{} vs. GPT-5 mini \model{} & 0.0291 & [-0.0072, 0.0728] & 1\\
Claude Sonnet 4.5 \model{} vs. Rule-only & 0.0776 & [0.0285, 0.1298] & 1\\
GPT-5 mini \model{} vs. Direct & 0.0921 & [-0.0083, 0.1669] & 7.41e-05\\
GPT-5 mini \model{} vs. Flat RAG & 0.1019 & [-0.0002, 0.1807] & 2.25e-05\\
GPT-5 mini \model{} vs. \model{} w/o verifier & -0.0246 & [-0.0531, -0.0053] & 1\\
GPT-5 mini \model{} vs. Rule-only & 0.0485 & [-0.0028, 0.0931] & 1\\\bottomrule\end{tabular}
\caption{Compositional: paired exploratory cross-backbone comparisons. Intervals use 10,000 entity-cluster bootstrap replicates; McNemar $p$ values are diagnostic and Holm-corrected within the 57-comparison family. Only the 42 completed comparisons are reported; correction retains the full family of 57 prespecified comparisons. Scenario-cluster sensitivity results are archived.}\label{tab:multi-stat-compositional_challenge}\end{table*}

\begin{table*}[tp]\centering\small\setlength{\tabcolsep}{3pt}\begin{tabular}{p{0.53\textwidth}ccc}\toprule Comparison & $\Delta$F1 & Entity 95\% CI & Holm $p$\\\midrule
Claude Opus 4.5 \model{} vs. Direct & 0.2630 & [0.2070, 0.3240] & 2.5e-22\\
Claude Opus 4.5 \model{} vs. Flat RAG & 0.2566 & [0.2059, 0.3131] & 4.67e-22\\
Claude Opus 4.5 \model{} vs. \model{} w/o verifier & 0.0000 & [0.0000, 0.0000] & 1\\
Claude Opus 4.5 \model{} vs. GPT-5 mini \model{} & 0.0304 & [-0.0177, 0.0869] & 1\\
Claude Opus 4.5 \model{} vs. Rule-only & 0.0032 & [-0.0372, 0.0480] & 0.0718\\
Claude Sonnet 4.5 \model{} vs. Direct & 0.2921 & [0.2217, 0.3707] & 2.14e-23\\
Claude Sonnet 4.5 \model{} vs. Flat RAG & 0.1886 & [0.1175, 0.2674] & 2.83e-16\\
Claude Sonnet 4.5 \model{} vs. \model{} w/o verifier & -0.0228 & [-0.0365, -0.0110] & 0.0083\\
Claude Sonnet 4.5 \model{} vs. GPT-5 mini \model{} & 0.0082 & [-0.0189, 0.0364] & 1\\
Claude Sonnet 4.5 \model{} vs. Rule-only & -0.0190 & [-0.0432, 0.0077] & 0.00436\\
GPT-5 mini \model{} vs. Direct & 0.0333 & [-0.0444, 0.1186] & 0.167\\
GPT-5 mini \model{} vs. Flat RAG & 0.0491 & [-0.0363, 0.1431] & 0.065\\
GPT-5 mini \model{} vs. \model{} w/o verifier & -0.0189 & [-0.0300, -0.0093] & 0.0322\\
GPT-5 mini \model{} vs. Rule-only & -0.0271 & [-0.0525, -0.0003] & 0.00507\\\bottomrule\end{tabular}
\caption{Controlled stress: paired exploratory cross-backbone comparisons. Intervals use 10,000 entity-cluster bootstrap replicates; McNemar $p$ values are diagnostic and Holm-corrected within the 57-comparison family. Only the 42 completed comparisons are reported; correction retains the full family of 57 prespecified comparisons. Scenario-cluster sensitivity results are archived.}\label{tab:multi-stat-controlled_stress}\end{table*}

The analysis uses a separate family of 57 prespecified comparisons: in each of the
three partitions, \model{} versus Direct, Flat RAG, and \model{} w/o verifier within each of four backbones; each
additional \model{} versus the primary GPT-5-mini \model{}; and each \model{} versus the shared
deterministic Rule-only. Paired entity-cluster bootstrap uses 10,000 resamples;
source-scenario clustering is a sensitivity analysis. Candidate-level McNemar
tests use the full 57-slot Holm allocation. A comparison is performed
only when both complete partitions are available; otherwise it receives no reported $p$ value and is omitted from the tables. Unperformed slots remain
in the multiplicity allocation. Dependence between candidates
limits the interpretation of these tests; effect sizes and cluster intervals
remain primary. Development summaries retain every repetition and report
population standard deviation and worst-case recall only when all three
repetitions are complete. Incomplete repetitions preclude a stability summary.

\begin{table*}[tp]\centering\small\setlength{\tabcolsep}{3pt}\begin{tabular}{llcccc}\toprule Backbone & Partition & F1 mean $\pm$ SD & Recall mean $\pm$ SD & Min. recall & Max. FRR\\\midrule
\multirow{3}{*}{GPT-5 mini} & Base & 0.899 $\pm$ 0.024 & 0.833 $\pm$ 0.115 & 0.688 & 0.000\\
 & Comp. & 0.645 $\pm$ 0.016 & 0.889 $\pm$ 0.079 & 0.778 & 0.000\\
 & Stress & 0.949 $\pm$ 0.013 & 0.833 $\pm$ 0.027 & 0.800 & 0.000\\
\addlinespace
\multirow{3}{*}{Claude Opus 4.5} & Base & 0.952 $\pm$ 0.000 & 1.000 $\pm$ 0.000 & 1.000 & 0.000\\
 & Comp. & 0.667 $\pm$ 0.000 & 1.000 $\pm$ 0.000 & 1.000 & 0.000\\
 & Stress & 0.972 $\pm$ 0.004 & 0.889 $\pm$ 0.016 & 0.867 & 0.000\\
\addlinespace
\multirow{3}{*}{Claude Sonnet 4.5} & Base & 0.914 $\pm$ 0.015 & 0.812 $\pm$ 0.068 & 0.719 & 0.000\\
 & Comp. & 0.595 $\pm$ 0.011 & 0.667 $\pm$ 0.045 & 0.611 & 0.000\\
 & Stress & 0.923 $\pm$ 0.013 & 0.767 $\pm$ 0.027 & 0.733 & 0.000\\\bottomrule\end{tabular}\caption{\model{} development stability on the same 250-scenario subset across three repetitions. SD is population SD. Results include all three repetitions for each reported backbone, including runs with low recall.}\label{tab:multi-stability}\end{table*}

\paragraph{Cross-backbone results.}
Tables~\ref{tab:multi-base}, \ref{tab:multi-compositional_challenge}, and
\ref{tab:multi-controlled_stress} report the complete-partition results for
base, compositional, and stress tests, respectively.
The Claude backbones have complete results for all evaluated methods and
partitions. GPT-4o mini contributes Direct results on stress and Flat RAG
results on compositional and stress partitions; conclusions for this backbone are restricted to those evaluated configurations. The 42 available paired comparisons retain the full
57-comparison multiplicity correction.

Across completed Claude configurations, \model{} exceeds Direct and Flat RAG
with zero observed false recommendations under the policy references.
Tables~\ref{tab:multi-stat-base}, \ref{tab:multi-stat-compositional_challenge},
and \ref{tab:multi-stat-controlled_stress} quantify paired differences;
Rule-only remains competitive, particularly on Stress. Repetition variability
(Table~\ref{tab:multi-stability}) limits claims of consistently reliable coverage.

\paragraph{Recall retained after verification.}
The Sonnet results make the coverage cost explicit. Recommendation recall
falls from 0.86 to 0.80 on base, 0.94 to 0.90 on compositional cases, and 0.95
to 0.85 on stress cases, while precision remains 1.00. F1 also falls throughout, without a compensating reduction in unsupported
recommendations. Opus retains recall of 0.98, 1.00, and 0.83, respectively,
with and without verification. Its unchanged test decisions also do not
establish stability across repeated execution: development stress recall
varies across the three runs in Table~\ref{tab:multi-stability}.

\end{document}